\documentclass{article}

\PassOptionsToPackage{numbers, compress}{natbib}

\usepackage[final]{neurips_2019}

\usepackage[utf8]{inputenc} 
\usepackage[T1]{fontenc}    
\usepackage{hyperref}       
\usepackage{url}            
\usepackage{booktabs}       
\usepackage{amsfonts}       
\usepackage{nicefrac}       
\usepackage{microtype}      
\usepackage{amsmath}
\usepackage{graphicx}
\usepackage{cleveref}
\usepackage{subcaption}
\usepackage[usenames,dvipsnames]{xcolor}
\usepackage{enumitem}       
\usepackage{commath}
\usepackage{textcomp}
\usepackage{gensymb}

\usepackage{tikz}

\usepackage[acronym,smallcaps,nowarn,section,nogroupskip,nonumberlist]{glossaries}
\setkeys{glslink}{hyper=false}
\newacronym{RL}{rl}{Reinforcement Learning}
\newacronym{VIB}{vib}{Variational Information Bottleneck}
\newacronym{IB}{ib}{Information Bottleneck}
\newacronym{IBAC}{ibac}{Information Bottleneck Actor Critic}
\newacronym{SNI}{sni}{Selective Noise Injection}
\newacronym{PPO}{ppo}{Proximal Policy Optimization}
\newacronym[firstplural=Markov decision processes]{MDP}{mdp}{Markov decision process}
\newacronym[firstplural=partially observable Markov decision processes]{POMDP}{pomdp}{partially observable Markov decision process}

\newacronym{IBAC-SNI}{IBAC-SNI}{}
\glsunset{IBAC-SNI}
\newacronym{PG}{pg}{...}
\glsunset{PG}
\newacronym{AC}{ac}{...}
\glsunset{AC}
\newacronym{KL}{kl}{...}
\glsunset{KL}

\usepackage{amsthm}

\title{Generalization in Reinforcement Learning with Selective Noise Injection and Information Bottleneck}

\author{%
  Maximilian Igl \thanks{Work performed during an internship at Microsoft Research Cambridge}\\
  University of Oxford\\
  \And
  Kamil Ciosek\\
  Microsoft Research\\
  \And
  Yingzhen Li\\
  Microsoft Research\\
  \And
  Sebastian Tschiatschek\\
  Microsoft Research\\
  \And
  Cheng Zhang\\
  Microsoft Research\\
  \And
  Sam Devlin \thanks{Co-Senior Authors}\\
  Microsoft Research \\
  \And
  Katja Hofmann \footnotemark[\value{footnote}]\\
  Microsoft Research\\
  }

\begin{document}

\maketitle
\begin{abstract}
    The ability for policies to generalize to new environments is key to the broad application
    of RL agents.
    A promising approach to prevent an agent's policy from overfitting to a limited set of training
    environments is to apply regularization techniques originally developed for supervised learning.
    However, there are stark differences between supervised learning and RL.
    We discuss those differences and propose modifications to existing regularization techniques in
    order to better adapt them to RL.
    In particular, we focus on regularization techniques relying on the injection of noise into the
    learned function, a family that includes some of the most widely used approaches such as Dropout
    and Batch Normalization.
    To adapt them to RL, we propose \emph{Selective Noise Injection} (SNI), which maintains the 
    regularizing effect the injected noise has, while mitigating the adverse effects it has on the gradient quality.
    Furthermore, we demonstrate that the Information Bottleneck (IB) is a particularly well suited regularization technique for RL as it is effective in the low-data regime encountered early on in training RL agents.
    Combining the IB with SNI, we significantly outperform current state of the art results, including on 
    the recently proposed generalization benchmark \emph{Coinrun}. 
\end{abstract}

\section{Introduction}

Deep \gls{RL} has been used to successfully train policies with impressive performance on a range of challenging tasks, including Atari \citep{bellemare2013arcade,hessel2018rainbow,mnih2013playing},
continuous control \citep{peng2018sim,schulman2017proximal} and 
tasks with long-ranged temporal dependencies \citep{openai2018dota}.
In those settings, the challenge is to be able to successfully explore and learn policies
complex enough to solve the training tasks.
Consequently, the focus of these works was to improve the learning performance of agents in the
training environment and less
attention was being paid to generalization to testing environments.

However, being able to generalize is a key requirement for the broad application of autonomous agents.
Spurred by several recent works showing that most \gls{RL} agents overfit to the training environment
\citep{henderson2018deep,whiteson2011protecting,zhang2018dissection,zhang2018study,zhao2019investigating},
multiple benchmarks to evaluate the generalization capabilities of agents were proposed, typically by procedurally
generating or modifying levels in video games
\citep{chaplot2016transfer,cobbe2018quantifying,juliani2019obstacle,justesen2018illuminating,nichol2018gotta,zhang2018natural}. How to learn generalizable policies in these environments remains an open question, but early results have shown the use of regularization techniques (like weight decay,
dropout and batch normalization) established in the supervised learning paradigm can also be useful for \gls{RL} agents \citep{cobbe2018quantifying}. Our work builds on these results, but highlights two important differences between supervised learning
and \gls{RL} which need to be taken into account when regularizing agents.

First, because in \gls{RL} the \emph{training data depends on the model} and, consequently, the
regularization method, stochastic regularization techniques like Dropout or BatchNorm can have adverse
effects.
For example, injecting stochasticity into the policy can lead
to prematurely ending episodes, preventing the agent from observing future rewards.
Furthermore, stochastic regularization can destabilize training through the
learned \emph{critic} and off-policy importance weights. 
To mitigate those adverse effects and effectively apply stochastic regularization techniques to
\gls{RL}, we propose \gls{SNI}.
It selectively applies stochasticity only when it serves regularization and
otherwise computes the output of the regularized networks deterministically.
We focus our evaluation on Dropout and the \gls{VIB}, but the
proposed method is applicable to most forms of stochastic regularization.

A second difference between \gls{RL} and supervised learning is the \emph{non-stationarity of the
data-distribution} in \gls{RL}. 
Despite many \gls{RL} algorithms utilizing millions or even billions of observations, the diversity of states encountered early on in
training can be small, making it difficult to learn general features. 
While it remains an open question as to why deep neural networks generalize despite being able to
perfectly memorize the training data \citep{belkin2018reconciling,zhang2016understanding}, 
it has been shown that the optimal point on the worst-case generalization bound requires the model
to rely on a more compressed set of features the fewer data-points we have
\citep{shamir2010learning,tishby2015deep}.
Therefore, to bias our agent towards more general features even early on in training, we adapt the
\gls{IB} principle to an actor-critic agent, which we call 
\gls{IBAC}. In contrast to other regularization techniques, \gls{IBAC} directly incentivizes the
compression of input features, resulting in features that are more robust under a shifting data-distribution
and that enable better generalization to held-out test environments. 

We evaluate our proposed techniques using \gls{PPO}, an off-policy actor-critic algorithm, on two challenging generalization tasks, \emph{Multiroom} \citep{gym_minigrid} and
\emph{Coinrun} \citep{cobbe2018quantifying}. 
We show the benefits of both \gls{IBAC} and \gls{SNI} individually as well as in combination,
with the resulting \gls{IBAC}-\gls{SNI} significantly outperforming the previous state of the art results. 

\section{Background}
\label{sec:rl-background}

We consider having a distribution $q(m)$ of \glspl{MDP} $m\in
\mathcal{M}$, with $m$ being a tuple $(\mathcal{S}_m, \mathcal{A}, T_m, R_m, p_m)$ consisting of state-space $\mathcal{S}_m$, action-space $\mathcal{A}$,
transition distribution $T_m(s'|s,a)$, reward function $R_m(s,a)$ and initial state distribution $p_m(s_0)$ \citep{puterman2014markov}.
For training, we either assume unlimited access to $q(m)$ (like in \cref{sec:multiroom}, 
\emph{Multiroom}) or restrict ourselves to a fixed set of
training environments $M_{\text{train}}=\{m_1,\dots,m_n\}$, $m_i\sim q$ 
(like in \cref{sec:coinrun}, \emph{Coinrun}). 

The goal of the learning process is to find a policy $\pi_\theta(a|s)$, parameterized by $\theta$,
which maximizes the discounted expected reward: $ J(\pi_\theta) = \mathbb{E}_{q,\pi,T_m,p_m} \left[ \sum_{t=0}^T \gamma^t R_m(s_t,a_t) \right]$. Although any \gls{RL} method with an off-policy correction term could be used with our proposed method of \gls{SNI}, \gls{PPO} \citep{schulman2017proximal} has shown strong
performance and enables direct comparison with prior work \citep{cobbe2018quantifying}.
The actor-critic version of this algorithm collects trajectory data $\mathcal{D}_\tau$ using a rollout policy
$\pi^r_\theta(a_t|s_t)$ and subsequently optimizes a surrogate loss: 
\begin{equation}
    \label{eq:ppo-pg}
    L_{\gls{PPO}} = -\mathbb{E}_{\mathcal{D}_\tau} \left[ \min(c_t(\theta) A_t, \text{clip}(c_t(\theta), 1-\epsilon, 1+\epsilon) A_t) \right] 
\end{equation}
with $c_t(\theta) = \frac{\pi_\theta(a_t|s_t)}{\pi_\theta^r(a_t|s_t)}$ for $K$ epochs. 
The advantage $A_t$ is computed as in A2C \citep{mnih2016asynchronous}.
This is an efficient approximate trust region method
\citep{schulman2015trust}, optimizing a pessimistic lower bound of the objective function on the
collected data. It corresponds to estimating the gradient w.r.t the policy conservatively, since
moving $\pi_\theta$ further away from $\pi_\theta^r$, such that $c_t(\theta)$ moves 
outside a chosen range $[1-\epsilon, 1+\epsilon]$, is only taken into account if it decreases performance. 
Similarly, the value function loss minimizes an upper bound on the squared error: 
\begin{equation}
    \label{eq:ppo-v}
L_{\gls{PPO}}^{V} = \mathbb{E}_{\mathcal{D}_\tau} \left[\frac{1}{2}\max\left((V_\theta - V^T)^2, (V^r + \text{clip}(V_\theta - V^r, 1-\epsilon, 1+\epsilon) - V_{\text{target}})^2\right)  \right]
\end{equation}
with a bootstrapped value function target $V_{\text{target}}$ \citep{mnih2016asynchronous} and previous value function $V^r$.
The overall minimization objective is then: 
\begin{equation}
    \label{eq:ppo-total}
    L_t(\theta) = L_{\gls{PPO}} + \lambda_V L_{\gls{PPO}}^V - \lambda_H H[\pi_\theta]
\end{equation}
where $H[\cdot]$ denotes an entropy bonus to encourage exploration and prevent the policy to
collapse prematurely. In the following, we discuss regularization techniques 
that can be used to mitigate overfitting to the states and \glspl{MDP} so far seen during training.

\subsection{Regularization Techniques in Supervised Learning}
\label{sec:generalization}

In supervised learning, classifiers are often regularized using a variety of techniques to prevent
overfitting. Here, we briefly present several major approaches which we either utilize as baseline or
extend to \gls{RL} in \cref{sec:method}.

\emph{Weight decay}, also called L2 regularization, reduces the magnitude of the
weights $\theta$ by
adding an additional loss term $\lambda_w \frac{1}{2}\|\theta\|^2_2$. With a gradient
update of the form $\theta \leftarrow \theta - \alpha \nabla_\theta (L(\theta) +
\frac{\lambda_w}{2}\|\theta\|^2_2)$, this decays the weights 
in addition to optimizing $L(\theta)$, i.e. we have $\theta\leftarrow (1-\alpha \lambda_w) \theta - \alpha\nabla_\theta L(\theta)$.

\emph{Data augmentation} refers to changing or distorting the available input data to improve
generalization.
In this work, we use a modified version of cutout \citep{devries2017improved}, proposed by
\citep{cobbe2018quantifying}, in which a random number of rectangular areas in the input image is
filled by random colors.

\emph{Batch Normalization} \citep{hoffer2017train,ioffe2015batch} normalizes activations of
specified layers by estimating their mean and variance using the current mini-batch.
Estimating the batch statistics introduces noise which has been shown to help improve generalization
\citep{luo2018towards} in supervised learning.

Another widely used regularization technique for deep neural networks is \emph{Dropout} \citep{srivastava2014dropout}. Here, during
training, individual activations are randomly zeroed out with a fixed probability $p_d$. This serves to
prevent co-adaptation of neurons and can be applied to any layer inside the network. 
One common choice, which we are following in our architecture, is to apply it to the last hidden layer.

Lastly, we will briefly describe the \acrfull{VIB} \citep{alemi2016deep}, a deep variational
approximation to the \acrfull{IB} \citep{tishby2000information}.
While not typically used for regularization in deep supervised learning, we demonstrate in \cref{sec:experiments} that
our adaptation \gls{IBAC} shows strong performance in \gls{RL}.
Given a data distribution $p(X,Y)$, the learned model $p_\theta(y|x)$ is
regularized by inserting a stochastic latent variable $Z$ and minimizing the mutual
information between the input $X$ and $Z$, $I(X,Z)$, while maximizing the
predictive power of the latent variable, i.e. $I(Z,Y)$. 
The \gls{VIB} objective function is:
\begin{equation}
    \label{eq:vib}
    L_{\text{VIB}} = \mathbb{E}_{p(x,y),p_\theta(z|x)}\big[ -\log q_\theta(y|z) + \beta D_{KL}[p_\theta(z|x)\| q(z)] \big]
\end{equation}
where $p_\theta(z|x)$ is the encoder, $q_\theta(y|z)$ the decoder, $q(z)$ the approximated latent marginal often
fixed to a normal distribution $\mathcal{N}(0, I)$ and $\beta$ is a hyperparameter. 
For a normal distributed $p_\theta(z|x)$, \cref{eq:vib} can be optimized by gradient decent using the reparameterization trick \citep{kingma2013auto}.

\section{The Problem of Using Stochastic Regularization in RL}
\label{sec:stochastic}

We now take a closer look at a prototypical objective for training actor-critic
methods and highlight important differences to supervised learning. Based on those observations, we propose an
explanation for the finding that some stochastic optimization methods are less effective \citep{cobbe2018quantifying} or can even be
detrimental to performance when combined with other regularization techniques (see
\cref{sec:ap:coinrun}).

In supervised learning, the optimization objective takes a form similar to
$\max_\theta \mathbb{E}_{\mathcal{D}}\left[ \log {\color{blue} p_\theta(y|x)} \right]$,
where we highlight the model ${\color{blue} p_\theta(y|x)}$ to be updated in {\color{blue} blue}, $\mathcal{D}$ is the
available data and $\theta$ the parameters to be learned.
On the other hand, in \gls{RL} the objective for the actor is to maximize
$J(\pi_\theta)=\mathbb{E}_{\color{blue}\pi_\theta(a|s)}\left[ \sum_t \gamma^t R_m(s_t,a_t) \right]$,
where, for convencience, we drop $q$, $T_m$ and $p_m$ from the notation of the expectation.
Because now the learned distribution, ${\color{blue} \pi_\theta(a|s)}$, is part of 
data-generation, computing the gradients, as done in policy gradient methods, requires the
log-derivative trick. For the class of deep off-policy actor-critic methods we are experimentally
evaluating in this paper, one also typically uses the policy gradient theorem \citep{sutton2000policy}
and an estimated \emph{critic} $V_\theta(s)$ as baseline and for bootstrapping to reduce
the gradient variance.
Consequently, the gradient estimation becomes:
\begin{equation}
 \label{eq:rl-objective1}
    \nabla_\theta J({\pi_\theta}) = \mathbb{E}_{\color{RedOrange} \pi^r_\theta(a_t|s_t)}\left[ \sum_t^T \frac{\color{blue}\pi_\theta(a_t|s_t)}{{\color{RedOrange}\pi^r_\theta(a_t|s_t)}} \nabla_\theta {\color{blue}\log \pi_\theta(a_t|s_t)} (r_t + \gamma {\color{RedOrange} V_\theta(s_{t+1}) - V_\theta(s_t)}) \right]
\end{equation}
where we utilize a \emph{rollout policy} ${\color{RedOrange} \pi^r_\theta}$ to collect trajectories. It can deviate from $\color{blue}\pi_\theta$ but should be
similar to keep the off-policy correction term
${\nicefrac{\color{blue}\pi_\theta}{\color{RedOrange} \pi^r_\theta}}$ low variance. In
\cref{eq:rl-objective1}, only the term ${\color{blue} \pi_\theta(a_t|s_t)}$ is being updated
and we highlight in {\color{RedOrange} orange} all the additional influences of the learned policy
and critic on the gradient.  

Denoting by the superscript $\perp$ that $\color{RedOrange}V_\theta^\perp$ is assumed constant, we can write 
the optimization objective for the critic as
\begin{equation}
 \label{eq:rl-objective2}
    L_{\text{AC}}^V = \min_\theta \mathbb{E}_{\color{RedOrange} \pi^r_\theta(a_t|s_t)}\left[\left( \gamma {\color{RedOrange} V_\theta^\perp(s_{t+1})} + r_t - {\color{blue} V_\theta(s_t)}  \right)^2\right]
\end{equation}
From \cref{eq:rl-objective1,eq:rl-objective2} we can see that the injection of noise into the
computation of ${\color{RedOrange} \pi_\theta^r}$ and ${\color{RedOrange} V_\theta}$ can degrade
performance in several ways:
i) During rollouts using the rollout policy ${\color{RedOrange} \pi_\theta^r}$, it can lead to undesirable actions, potentially
ending episodes prematurely, and thereby deteriorating the quality of the observed data;
ii) It leads to a higher variance of the off-policy correction term $
\nicefrac{\color{blue}\pi_\theta}{\color{RedOrange}\pi^r_\theta}$ because the
injected noise can be different for $\color{blue}\pi_\theta$ and $\color{RedOrange} \pi_\theta^r$,
increasing gradient variance;
iii) It increases variance in the gradient updates of both the policy and the critic
through variance in the computation of ${\color{RedOrange} V_\theta}$.

\section{Method}
\label{sec:method}
To utilize the strength of noise-injecting regularization techniques in \gls{RL}, we introduce      
\acrfull{SNI} in the following section. Its goal is to allow us to make use of such techniques while
mitigating the adverse effects the added stochasticity can have on the \gls{RL} gradient
computation. Then, in \cref{sec:ibac}, we propose \acrfull{IBAC} as a new regularization method and 
detail how \gls{SNI} applies to \gls{IBAC}, resulting in our state-of-the art method \gls{IBAC}-\gls{SNI}.

\subsection{Selective Noise Injection}

We have identified three sources of negative effects due to noise which we need to mitigate: In the rollout policy ${\color{RedOrange}\pi^r_\theta}$, in the critic ${\color{RedOrange}V_\theta}$ and in the off-policy correction term
$\nicefrac{\color{blue}\pi_\theta}{\color{RedOrange}\pi^r_\theta}$. We first introduce a short notation for \cref{eq:rl-objective1} as $ \nabla_\theta J(\pi_\theta) = \mathcal{G}_{\gls{AC}}({\color{RedOrange}\pi^r_\theta},
{\color{blue}\pi_\theta}, {\color{RedOrange}V_\theta})$. 

To apply \gls{SNI} to a regularization technique relying on noise-injection, we need to be able to
\emph{temporarily suspend} the noise and compute the output of the model deterministically. This is possible for
most techniques\footnote{In this work, we will focus on \gls{VIB} and Dropout as those show the most
promising results without \gls{SNI} (see \cref{sec:experiments}) and will leave its application to
other regularization techniques for future work.}: For example, in Dropout, we can freeze one particular dropout mask, 
in \gls{VIB} we can pass in the mode instead of sampling from the posterior
distribution and in Batch Normalization we can either utilize the moving average instead of the batch
statistics or freeze and re-use one statistic multiple times. Formally, we denote by
$\bar{\pi}_\theta$ the version of a component $\pi_\theta$, with the injected regularization noise suspended. Note
that this does not mean that $\bar{\pi}_\theta$ is deterministic, for example when the network approximates
the parameters of a distribution. 

Then, for \gls{SNI} we
modify the policy gradient loss as follows: i) We use $\color{RedOrange}\bar{V}_\theta$ as critic 
instead of $\color{RedOrange}V_\theta$ in both \cref{eq:rl-objective1,eq:rl-objective2}, eliminating unnecessary
noise through the critic; 
ii) We use $\color{RedOrange}\bar{\pi}^r$ as rollout policy instead of $\color{RedOrange}\pi^r$. For some regularization
techniques this will reduce the probability of undesirable actions; iii) We compute the policy gradient as a
\emph{mixture} between gradients for $\color{blue}\pi_\theta$ and $\color{blue}\bar{\pi}_\theta$ as follows:
\begin{equation}
    \label{eq:sni-pg}
    \mathcal{G}^{\gls{SNI}}_{\gls{AC}}({\color{RedOrange}\pi_\theta^r}, {\color{blue}\pi_\theta}, {\color{RedOrange}V_\theta}) = 
    \lambda \mathcal{G}_{\gls{AC}}({\color{RedOrange}\bar{\pi}_\theta^r}, {\color{blue}\bar{\pi}_\theta}, {\color{RedOrange}\bar{V}_\theta})  + 
    (1-\lambda) \mathcal{G}_{\gls{AC}}({\color{RedOrange}\bar{\pi}_\theta^r}, {\color{blue}\pi_\theta}, {\color{RedOrange}\bar{V}_\theta})
\end{equation}

The first term guarantees a lower variance of the off-policy importance weight, which is especially important early
on in training when the network has not yet learned to compensate for the injected noise.
The second term uses the noise-injected policy for updates, thereby taking advantage of its
regularizing effects while still reducing unnecessary variance through the use of $\color{RedOrange}\bar{\pi}^r$ and
$\color{RedOrange}\bar{V}_\theta$. 
Note that sharing the rollout policy $\color{RedOrange}\bar{\pi}^r$ between both terms allows us to use the
same collected data. Furthermore most computations are shared between both terms or can be
parallelized.

\subsection{Information Bottleneck Actor Critic}
\label{sec:ibac}

Early on in training an \gls{RL} agent, we are often faced with little variation in the training
data. Observed states are distributed only around the initial states $s_0$, making spurious correlations in the low
amount of data more likely. Furthermore, because neither the policy nor the critic have
sufficiently converged yet, we have a high variance in the target values of our loss function.

This combination makes it harder and less likely for the network to learn desirable features that are robust
under a shifting data-distribution during training and generalize well to held-out test
\glspl{MDP}.
To counteract this reduced signal-to-noise ratio, our goal is to explicitly bias the learning
towards finding more \emph{compressed} features which are shown to have a tighter worst-case
generalization bound \citep{tishby2015deep}. 
While a higher compression does not guarantee robustness under a \emph{shifting} data-distribution, we
believe this to be a reasonable assumption in the majority of \glspl{MDP}, for example because they
rely on a consistent underlying transition mechanism like physical laws.

To incentivize more compressed features, we use an approach similar to the \gls{VIB} \citep{alemi2016deep}, 
which minimizes the mutual information $I(S,Z)$ between the
state $S$ and its latent representation $Z$ while maximizing $I(Z,A)$, the predictive power of $Z$ on actions $A$.
To do so, we re-interpret the policy gradient update as maximization of the log-marginal likelihood
of $\pi_\theta(a|s)$ under the data distribution $p(s,a):=\frac{\rho^\pi(s)\pi_\theta(a|s)A^\pi(s,a)}{\mathcal{Z}}$
with discounted state distribution $\rho^\pi(s)$, advantage function $A^\pi(s,a)$ and normalization
constant $\mathcal{Z}$. Taking the semi-gradient of this objective, i.e. assuming $p(s,a)$ to be fixed, recovers the policy gradient:
\begin{equation}
    \nabla_\theta \mathcal{Z} \, \mathbb{E}_{p(s,a)}[\log \pi_\theta(a|s)] = \int \rho^\pi(s) \pi_\theta(a|s) \nabla_\theta \log \pi_\theta(a|s) A^\pi(s,a) \dif s \dif a.
\end{equation}
Now, following the same steps as \citep{alemi2016deep}, we introduce a stochastic latent
variable $z$ and minimize $\beta I(S,Z)$ while maximizing $I(Z,A)$ under $p(s,a)$, resulting in the
new objective:
\begin{equation}
    \begin{split}
    L_{\gls{IB}} & = \mathbb{E}_{p(s,a),p_\theta(z|s)}\big[ -\log q_\theta(a|z) + \beta D_{KL}[p_\theta(z|s)\| q(z)] \big] 
    \end{split}
\end{equation}
We take the gradient and use the reparameterization trick \citep{kingma2013auto} to write the encoder
$p_\theta(z|s)$ as deterministic function $f_\theta(s, \epsilon)$ with $\epsilon \sim p(\epsilon)$:
\begin{equation}
    \begin{split}
        \nabla_\theta L_{\gls{IB}} & = - \mathbb{E}_{\rho^\pi(s) \pi_\theta(a|s) p(\epsilon)} \left[ \nabla_\theta \log q_\theta(a|f_\theta(s, \epsilon)) A^\pi(s,a)\right] + \nabla_\theta \beta D_{KL}[p_\theta(z|s)\| q(z)] \\
        & = \nabla_\theta( L^{\gls{IB}}_{\gls{AC}} + \beta L^{\gls{KL}}),
    \end{split}
\end{equation}
resulting in a modified policy gradient objective and an additional regularization term $L^{\gls{KL}}$.

Policy gradient algorithms heuristically add an entropy bonus $H[\pi_\theta(a|s)]$ to prevent the policy
distribution from collapsing. However, this term also influences the distributions over $z$.
In practice, we are only interested in preventing $q_\theta(a|z)$ (not $\pi_\theta(a|s)=\mathbb{E}_z[q_\theta(a|z)]$)
from collapsing because our rollout policy $\bar{\pi}_\theta$ will not rely on stochasticity in $z$.
Additionally, $p_\theta(z|s)$ is already entropy-regularized by the \gls{IB} loss term\footnote{We have
$D_{\gls{KL}}[p_\theta(z|s)\| r(z)] = \mathbb{E}_{p_\theta(z|s)}[\log p_\theta(z|s) - \log r(z)] = -
H[p_\theta(z|s)] - \mathbb{E}_{p_\theta(z|s)}[\log r(z)]$}.
Consequently, we adapt the heuristic entropy bonus to
\begin{equation}
H^{\gls{IB}}[\pi_\theta(a|s)] := \int p_\theta(s, z) H[q_\theta(a|z)] \dif s \dif z,
\end{equation}
resulting in the overall loss function of the proposed \acrfull{IBAC}
\begin{equation}
   L^{\gls{IBAC}}_t(\theta) = L_{\gls{AC}}^{\gls{IB}} + \lambda_V L_{\gls{AC}}^V - \lambda_H H^{\gls{IB}}[\pi_\theta] + \beta L^{\gls{KL}}
\end{equation}
with the hyperparameters $\lambda_V$, $\lambda_H$ and $\beta$ balancing the loss terms.

While \gls{IBAC} incentivizes more compressed features, it also introduces stochasticity.
Consequently, combining it with \gls{SNI} improves performance, as we demonstrate in
\cref{sec:multiroom,sec:coinrun}.
To compute the noise-suspended policy $\bar{\pi}_\theta$ and critic $\bar{V}_\theta$, we use the
mode $z=\mu_\theta(s)$ as input to $q_\theta(a|z)$ and $V_\theta(z)$, 
where $\mu_\theta(s)$ is the mode of $p_\theta(z|s)$ and $V_\theta(z)$ now conditions on $z$ instead
of $s$, also using the compressed features.
Note that for \gls{SNI} with $\lambda=1$, i.e. with only the term
$\mathcal{G}_{\gls{AC}}(\bar{\pi}_\theta^r, \bar{\pi}_\theta, \bar{V}_\theta)$, this effectively
recovers a L2 penalty on the activations since the variance of $z$ will then always be ignored and the KL-divergence
between two Gaussians minimizes the squared difference of their means.

\section{Experiments}
\label{sec:experiments}

In the following, we present a series of experiments to show that the \gls{IB} finds more general features in the low-data regime and that this translates to improved generalization in \gls{RL} for \gls{IBAC} agents, especially when combined with \gls{SNI}. 
We evaluate our proposed regularization techniques on two environments, one grid-world with challenging
generalization requirements \citep{gym_minigrid} in which most previous approaches are unable to find the solution and 
on the recently proposed \emph{Coinrun} benchmark \citep{cobbe2018quantifying}. 
We show that \gls{IBAC}-\gls{SNI} outperforms previous state of the art on both environments
by a large margin.
Details about the used hyperparameters and network architectures can be found in the Appendix, code
to reproduce the results can be found at \url{https://github.com/microsoft/IBAC-SNI/}.

\subsection{Learning Features in the Low-Data Regime}
\label{sec:toy}

\begin{figure}[ht]
    \centering
    \begin{subfigure}{0.4\columnwidth}
        \includegraphics[width=\linewidth]{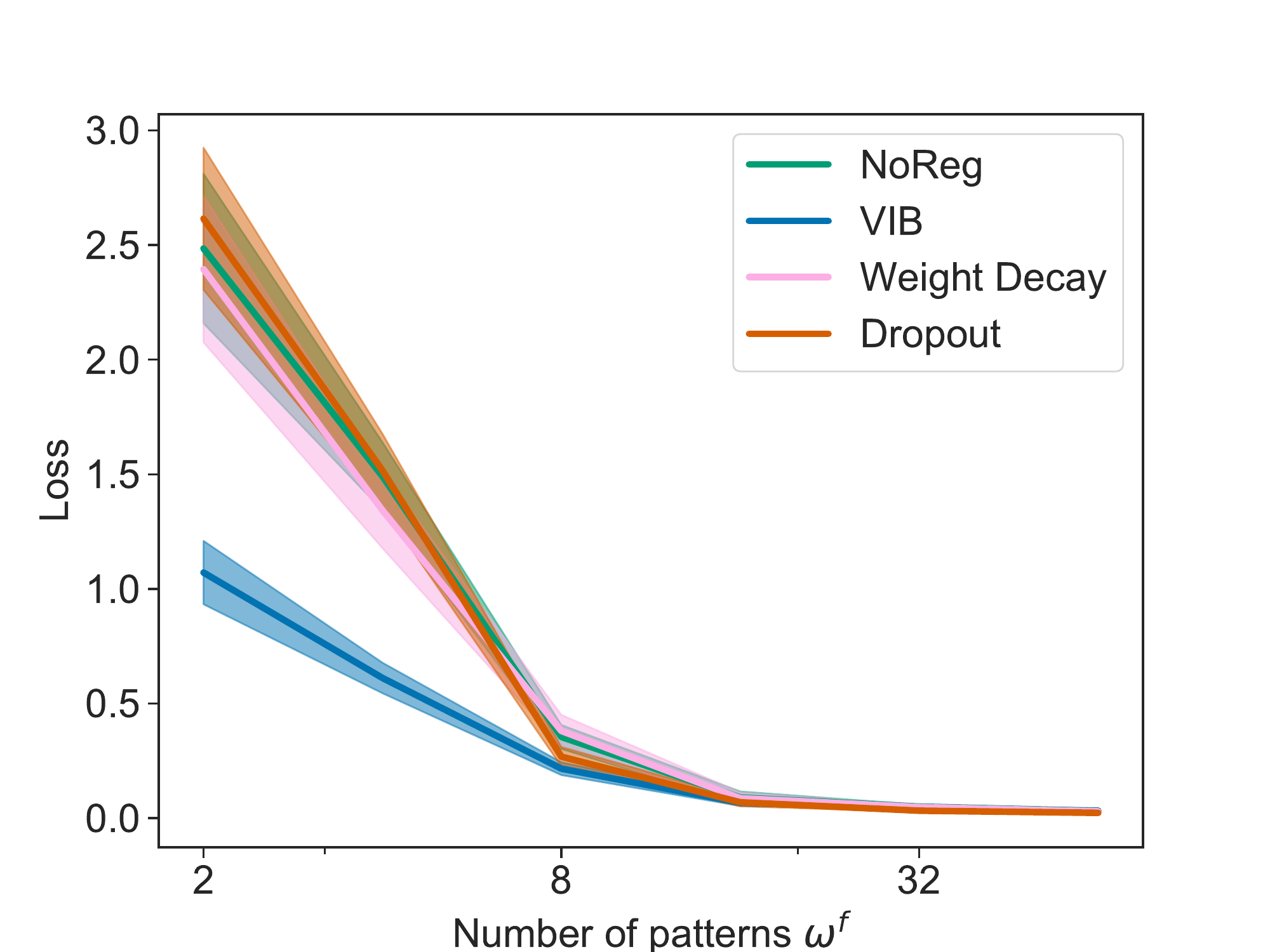}
    \end{subfigure}
    \begin{subfigure}{0.4\columnwidth}
        \includegraphics[width=\linewidth]{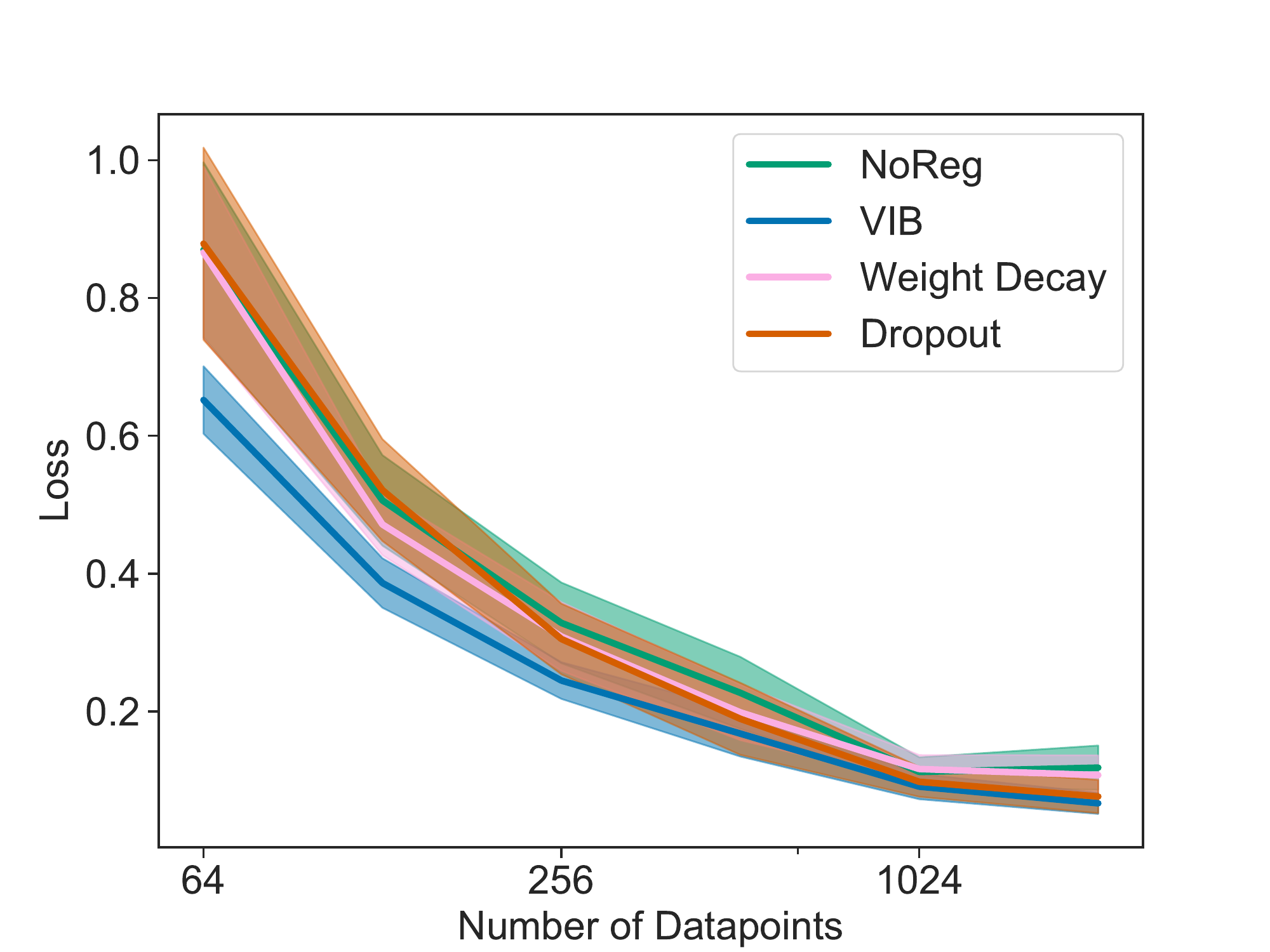}
    \end{subfigure}
    \caption{We show the loss on the test-data (lower is better). \emph{Left:} Higher $\omega^f$
    result in a larger difference in generality between features $f^c$ and $g^c$, making it easier
    to fit to the more general $g^c$. \emph{Right:} Learning $g^c$ with fewer datapoints is more challenging, but needed early in training \gls{RL} agents.}
    \label{fig:toy}
\end{figure}

First we start in the supervised setting and show on a synthetic dataset that the \gls{VIB} is particularly strong at finding more general features in the
low-data regime and in the presence of multiple signals with varying degrees of generality.
Our motivation is that the low-data regime is commonly encountered in
\gls{RL} early on in training and many environments allow the agent to base its decision
on a variety of features in the state, of which we would like to find the most general ones.

We generate the training dataset $\mathcal{D}_{\text{train}}=\{(c_i,x_i)\}_{i=1}^N$ with observations $x_i\in\mathbb{R}^{d_x}$ and classes $c_i\in\{1,\dots, n_c\}$. Each data point $i$ is
generated by first drawing the class $c_i\sim\mathit{Cat}(n_c)$ from a uniform categorical
distribution and generating the vector $x_i$ by embedding the information about $c_i$ in \emph{two}
different ways $g^c$ and $f^c$ (see \cref{sec:ap:toy} for details).
Importantly, only $g^c$ is shared between the training and test set. This allows us to measure the
model's relative reliance on $g^c$ and $f^c$ by measuring the test performance (all models perfectly
fit the training data).
We allow $f^c$ to encode the information about $c_i$ in $\omega^f$ different ways. Consequently, the higher $\omega^f$, the less general $f^c$ is.

In \cref{fig:toy} we measure how the test performance of fully trained classification models
varies for different regularization techniques
when we i) vary the generality of $f^c$ and ii) vary the number of data-points in the training set.
We find that most techniques perform comparably with the exception of the \gls{VIB} which is able to
find more general features both in the low-data regime and in the presence of multiple features with
only small differences in generality. In the next section, we show that this translates to faster
training and performance gains in \gls{RL} for our proposed algorithm \gls{IBAC}.

\subsection{Multiroom}
\label{sec:multiroom}

\begin{figure}[ht]
    \centering
    \begin{subfigure}{0.3\columnwidth}
        \centering
        \includegraphics[width=0.7\linewidth]{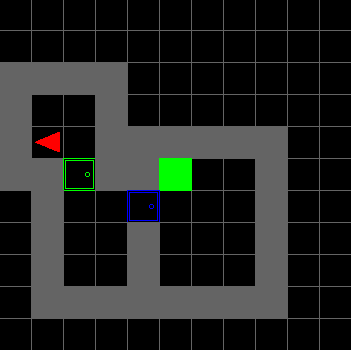}
    \end{subfigure}
    \begin{subfigure}{0.32\columnwidth}
        \includegraphics[width=\linewidth]{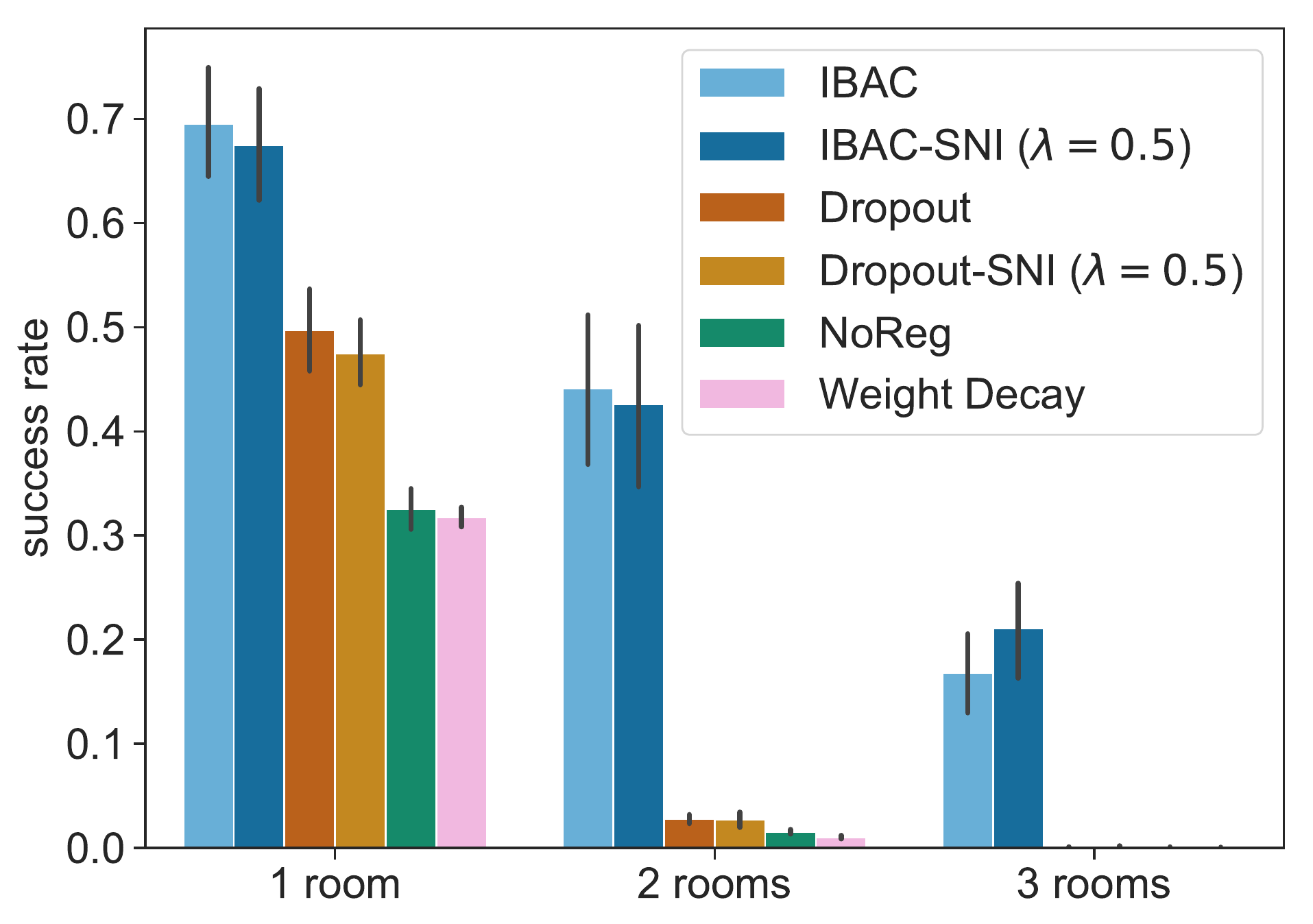}
    \end{subfigure}
    \begin{subfigure}{0.32\columnwidth}
        \includegraphics[width=\linewidth]{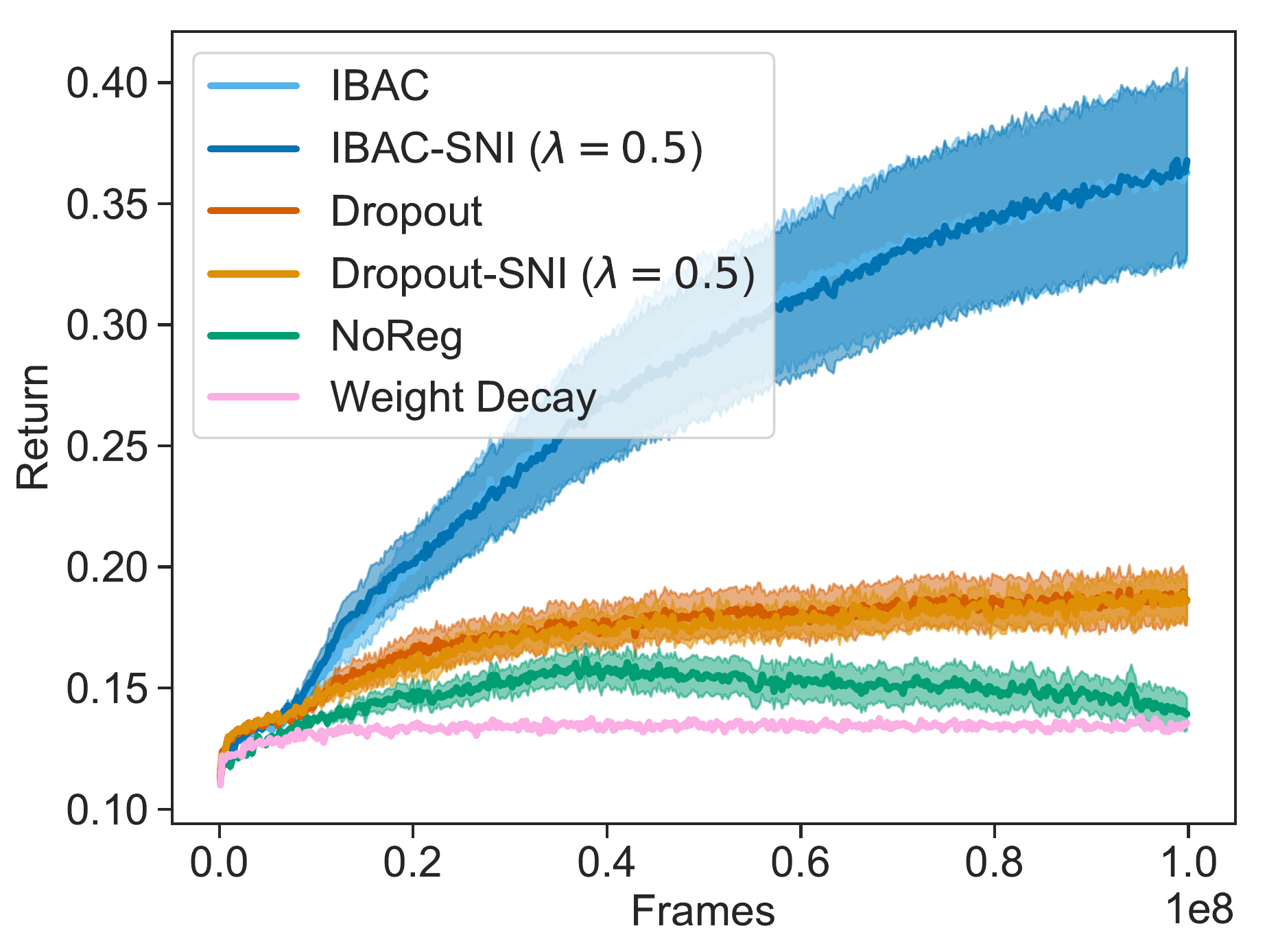}
    \end{subfigure}
    \caption{\emph{Left:} Typical layout of the environment. The red triangle denotes the agent and
    its direction, the green full square is the goal, colored boxes are doors and grey squares
    are walls.
    \emph{Middle:} Probability of finding the goal depending on level size for models trained on all
    levels. Shown are mean and standard error across 30 different seeds.
    \emph{Right:} Mean and standard error over of the return of the same models averaged across all room sizes.}
    \label{fig:grid}
\end{figure}

In this section, we show how \gls{IBAC} can help learning in \gls{RL} tasks which require
generalization. For this task, we do not distinguish between training and testing, but for each
episode, we draw $m$ randomly from the full distribution over \glspl{MDP} $q(m)$. As the number of \glspl{MDP} is
very large, learning can only be successful if the agent learns general features that are
transferrable between episodes.

This experiment is based on \citep{gym_minigrid}. The aim of the agent is to traverse a sequence of rooms to reach the goal (green square in
\cref{fig:grid}) as quickly as possible. It takes discrete actions to rotate $90\degree$ in either direction, move forward and toggle
doors to be open or closed. The observation received by the agent includes the full grid, one pixel
per square, with object type and object status (like direction) encoded in the 3 color channels.
Crucially, for each episode, the layout is generated randomly by placing a random number of rooms
$n_r\in\{1, 2, 3\}$ in a sequence connected by one door each.

The results in \cref{fig:grid} show that \gls{IBAC} agents are much better at
successfully learning to solve this task, especially for layouts with more rooms. 
While all other fully trained agents can solve less than 3\% of the layouts with two rooms and none of the ones
with three, \gls{IBAC}-\gls{SNI} still succeeds in an impressive 43\% and 21\% of those layouts.
The difficulty of this seemingly simple task arises from its generalization requirements: 
Since the layout is randomly generated in each episode, each state is observed very rarely,
especially for multi-room layouts, requiring generalization to allow learning.  
While in the 1 room layout the reduced policy stochasticity of the \gls{SNI} agent slightly reduces
performance, it improves performance for more complex layouts in which higher noise becomes detrimental.
In the next section we will see that this also holds for the much more complex \emph{Coinrun}
environment in which \gls{SNI} significantly improves the \gls{IBAC} performance.

\subsection{Coinrun}
\label{sec:coinrun}

\begin{figure}[ht]
    \begin{subfigure}[b]{0.3\linewidth}
    \includegraphics[width=\linewidth]{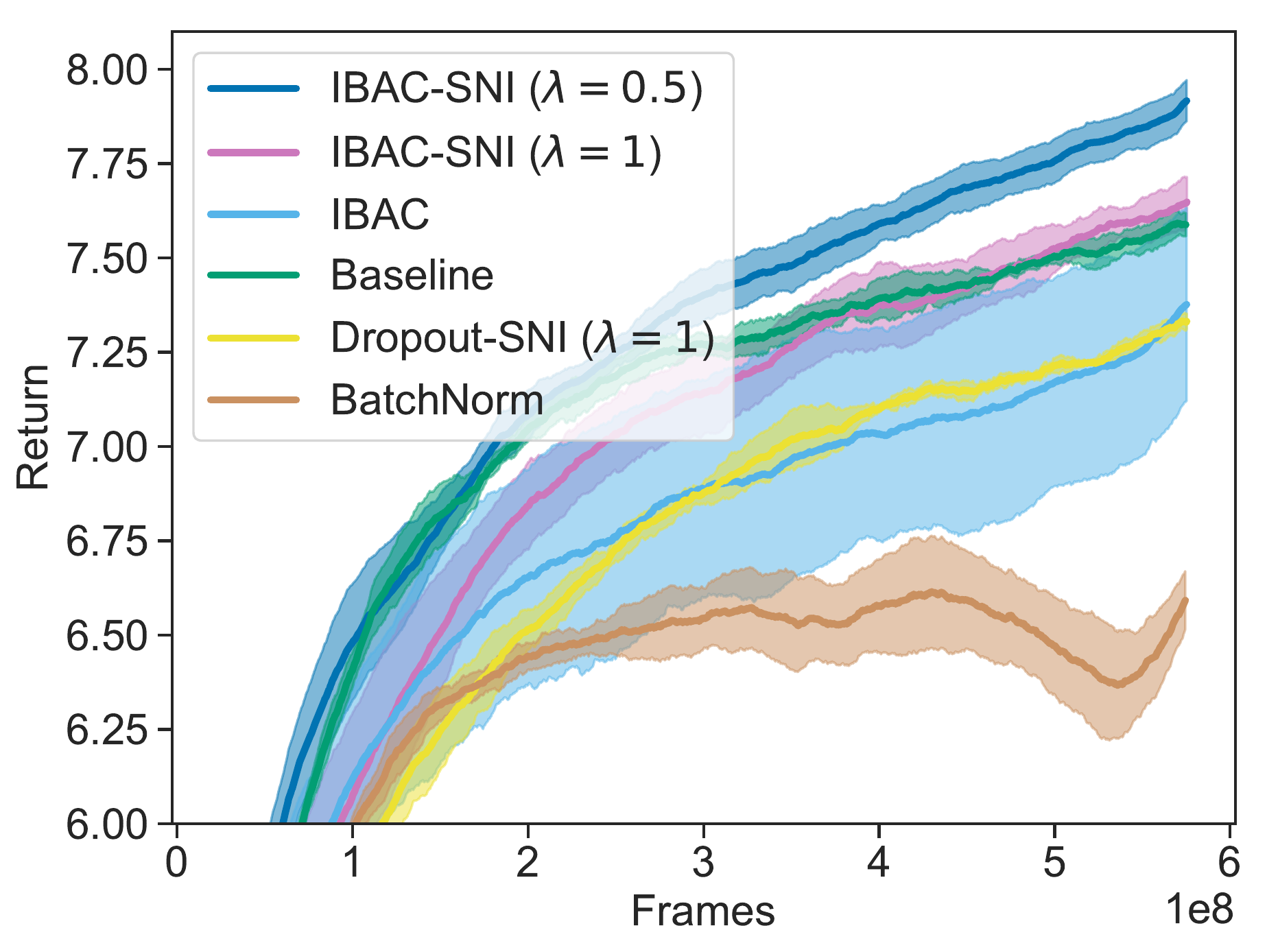}
    \end{subfigure}
    \centering
    \begin{subfigure}[b]{0.3\linewidth}
    \includegraphics[width=\linewidth]{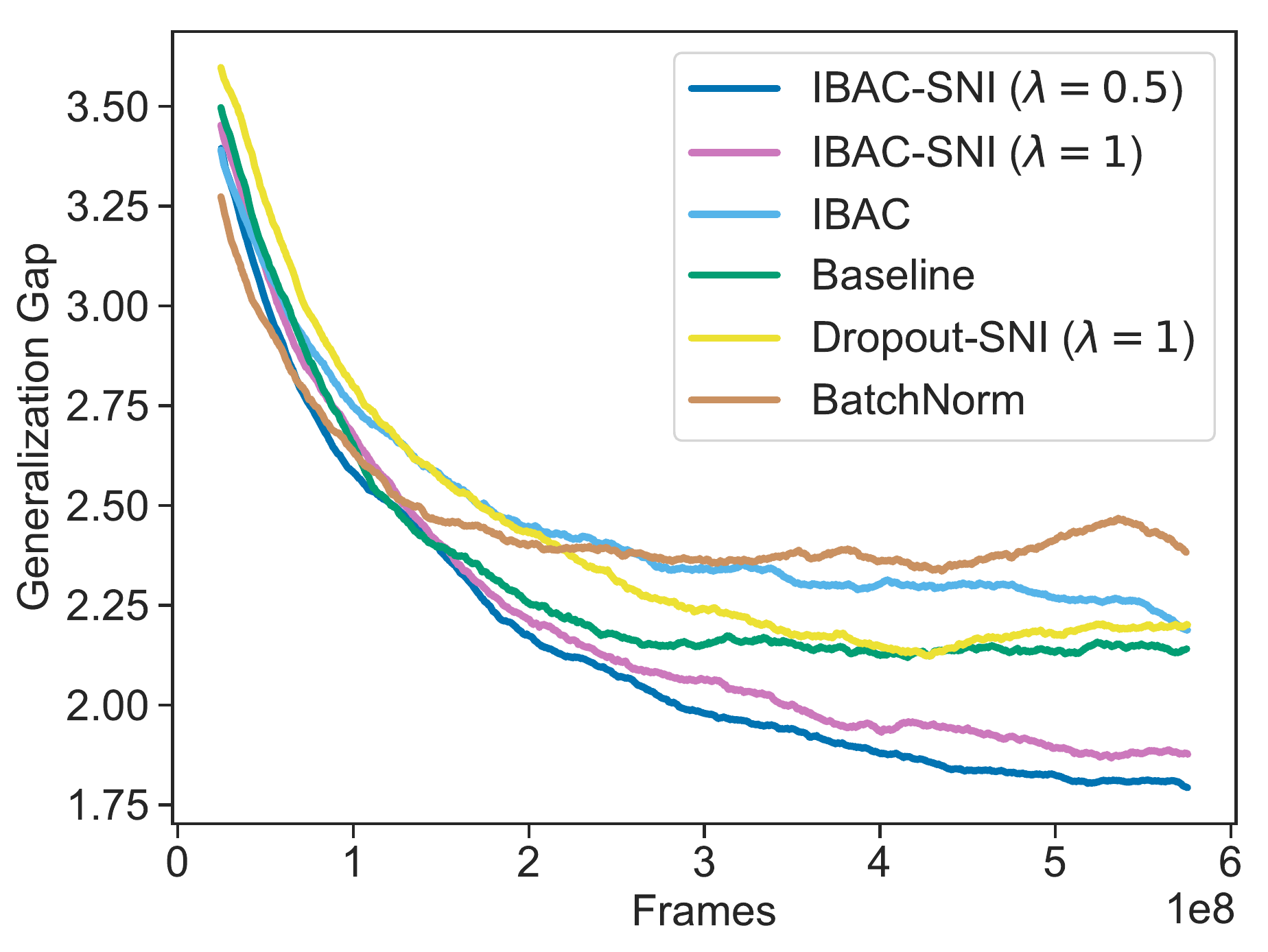}
    \end{subfigure}
    \centering
    \begin{subfigure}[b]{0.3\columnwidth}
    \includegraphics[width=\linewidth]{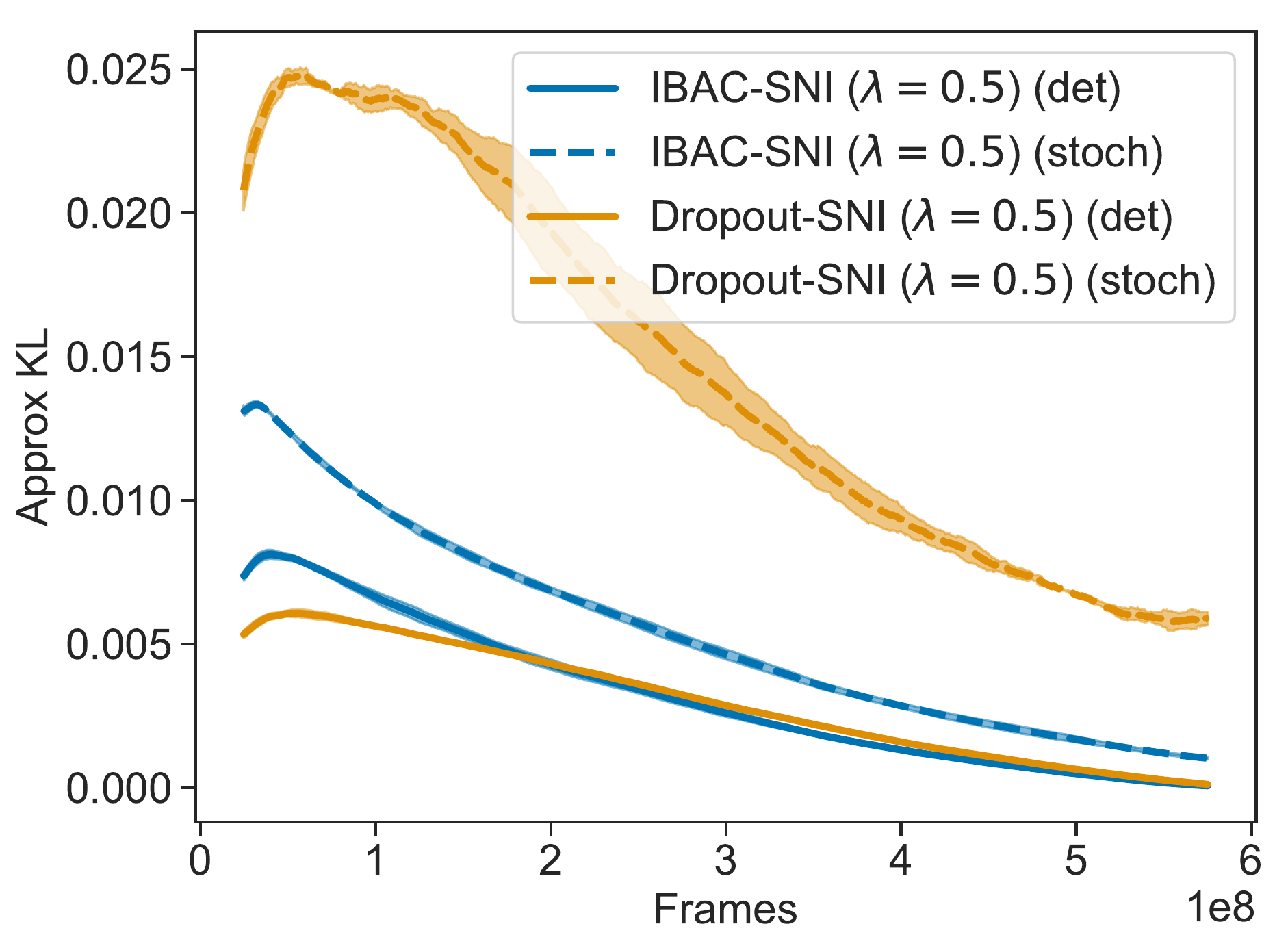}
    \end{subfigure}
    \caption{\emph{Left:} Performance of various agents on the test environments. We note that
    `BatchNorm' corresponds to the best performing agent in \citep{cobbe2018quantifying}. Furthermore,
    `Dropout-\gls{SNI} ($\lambda=1$)' is similar to the Dropout implementation used in
    \citep{cobbe2018quantifying} but was previously not evaluated with weight decay and data
    augmentation.
    \emph{Middle:} Difference between test performance and train performance (see
    \cref{fig:ap:coinrun1}). Without standard deviation for readability.
    \emph{Right:} Averaged approximate KL-Divergence between rollout policy and updated policy, used
    as proxy for the variance of the importance weight.
    Mean and standard deviation are across three random seeds.
    }
    \label{fig:coinrun}
\end{figure}

On the previous environment, we were able to show that \gls{IBAC} and \gls{SNI} help agents to find
more general features and to do so faster. Next, we show that this can lead to a higher final
performance on previously unseen test environments.
We evaluate our proposed regularization techniques on \emph{Coinrun} \citep{cobbe2018quantifying}, a
recently proposed generalization benchmark with high-dimensional observations and a large variety in
levels.
Several regularization techniques were previously evaluated there, making it an ideal
evaluation environment for \gls{IBAC} and \gls{SNI}.
We follow the setting proposed in \citep{cobbe2018quantifying}, using
the same 500 levels for training and evaluate on randomly drawn, new
levels of only the highest difficulty.

As \citep{cobbe2018quantifying} have
shown, combining multiple regularization techniques can improve performance, with their best-
performing agent utilizing data augmentation, weight decay and batch normalization.
As our goal is to push the state of the art on this environment and to accurately compare against their
results, \cref{fig:coinrun} uses weight decay and data-augmentation on all experiments.
Consequently, `Baseline' in \cref{fig:coinrun} refers to \emph{only} using weight decay and
data-augmentation whereas the other experiments use Dropout, Batch Normalization or \gls{IBAC}
\emph{in addition} to weight decay and data-augmentation. Results without those baseline techniques
can be found in \cref{sec:ap:coinrun}.

First, we find that almost all previously proposed regularization techniques \emph{decrease} performance compared to
the baseline, see \cref{fig:coinrun} (left), with batch normalization performing worst, possibly due to its unusual interaction with weight decay
\citep{laarhoven2017l2}. Note that this combination with batch normalization was the highest
performing agent in \citep{cobbe2018quantifying}. 
We conjecture that regularization techniques relying on stochasticity can introduce additional
instability into the training update, possibly deteriorating performance, especially if their
regularizing effect is not sufficiently different from what weight decay and data-augmentation
already achieve. This result applies to both batch normalization and Dropout, with and without
\gls{SNI}, although \gls{SNI} mitigates the adverse effects. 
Consequently, we can already improve on the state of the art by only relying on those two non-stochastic techniques. Furthermore, we find that \gls{IBAC} in combination with \gls{SNI} is able to significantly outperform
our new state of the art baseline. We also find that for
\gls{IBAC}, $\lambda=0.5$ achieves better performance than $\lambda=1$, justifying using both terms
in \cref{eq:sni-pg}. 

As a proxy for the variance of the off-policy correction term $\nicefrac{\pi^r_\theta}{\pi_\theta}$,
we show in \cref{fig:coinrun} (right) the estimated, averaged KL-divergence between the rollout
policy and
the update policy for both terms, $\mathcal{G}_{\gls{AC}}(\bar{\pi}_\theta^r, \bar{\pi}_\theta,
\bar{V}_\theta)$, denoted by `(det)' and $\mathcal{G}_{\gls{AC}}(\bar{\pi}_\theta^r, \pi_\theta,
\bar{V}_\theta)$, denoted by `(stoch)'. Because \gls{PPO} uses data-points multiple times it is
non-zero even for the deterministic term.
First, we can see that using the deterministic version reduces the KL-Divergence, explaining the
positive influence of $\mathcal{G}_{\gls{AC}}(\bar{\pi}_\theta^r, \bar{\pi}_\theta, \bar{V}_\theta)$. Second,
we see that the KL-Divergence of the stochastic part is much higher for Dropout than for \gls{IBAC},
offering an explanation of why for Dropout relying on purely the deterministic part ($\lambda=1$)
outperforms an equal mixing $\lambda=0.5$ (see \cref{fig:ap:coinrun2}). 

\section{Related Work}
\label{sec:related_work}

Generalization in RL can take a variety of forms, each necessitating
different types of regularization.
To position this work, we distinguished two types that, whilst not mutually exclusive, we believe to be conceptually distinct and found useful to isolate when studying approaches to improve generalization. 

The first type, \emph{robustness to uncertainty} refers to settings in which the 
unobserved \gls{MDP} $m$ influences the transition dynamics or reward structure. 
Consequently the current state $s$ might not contain enough information to act optimally in the current \gls{MDP}
and we need to find the action which is optimal under the uncertainty about $m$.
This setting often arises in robotics and control where exact physical
characteristics are unknown and domain shifts can occur \citep{levine2018learning}.
Consequently, \emph{domain randomization}, the injection of randomness into the environment, is often purposefully applied during training to allow for
sim-to-real transfer \citep{koos2013transferability,tobin2017domain}. Noise can be injected into the
states of the environment \citep{stulp2011learning} or the parameters of the transition distribution
like friction coefficients or mass values 
\citep{antonova2017reinforcement,packer2018assessing,yu2017preparing}.
The noise injected into the dynamics can also be manipulated
adversarially \citep{mandlekar2017adversarially,pinto2017robust,rajeswaran2016epopt}.
As the goal is to prevent overfitting to specific \glspl{MDP}, it also has been found that using
smaller \citep{rajeswaran2017towards} or simpler \citep{zhao2019investigating} networks can help.
We can also aim to learn an adaptive policy by treating the environment as \gls{POMDP}
\citep{peng2018sim,yu2017preparing} (similar to viewing the learning problem in the framework of Bayesian RL
\citep{poupart2006analytic}) or as a meta-learning problem \citep{al2017continuous,clavera2018learning,duan2016rl,schaul2015universal,sung2017learning,wang2016learning}.

On the other hand, we distinguish \emph{feature robustness}, which applies to environments with
high-dimensional observations (like images) in which generalization to previously unseen states can
be improved by learning to extract better features, as the focus for this paper. 
Recently, a range of benchmarks, typically utilizing procedurally generated levels, have been
proposed to evaluate this type of generalization
\citep{beeching2019deep,cobbe2018quantifying,johnson2016malmo,juliani2019obstacle,justesen2018illuminating,kanagawa2019rogue,nichol2018gotta,wydmuch2018vizdoom,zhang2018natural}.

Improving generalization in those settings can rely on generating more diverse observation data
\citep{cobbe2018quantifying,sadeghi2016cad2rl,tobin2017domain}, or strong, often relational,
inductive biases applied to the architecture
\citep{kansky2017schema,srouji2018structured,zambaldi2018deep}.  
Contrary to the results in continuous control domains, here 
deeper networks have been found to be more successful \citep{brutzkus2018why,cobbe2018quantifying}.
Furthermore, this setting is more similar to that of supervised learning, so
established regularization techniques like weight decay, dropout or batch-normalization have also successfully been
applied, especially in settings with a limited number of training environments \citep{cobbe2018quantifying}.
This is the work most closely related to ours.
We build on those results and improve upon them by taking into account the specific ways  in which \gls{RL} is \emph{different} from the supervised setting. They also do not consider the \gls{VIB} as a regularization technique. 

Combining \gls{RL} and \gls{VIB} has been recently explored for learning goal-conditioned policies
\cite{goyal2019transfer} and meta-RL \cite{rakelly2019efficient}. Both of these previous works
\cite{goyal2019transfer,rakelly2019efficient} also differ from the \gls{IBAC} architecture we
propose by conditioning action selection on both the encoded and raw state observation. These
studies complement the contribution made here by providing evidence that the \gls{VIB} can be used
with a wider range of \gls{RL} algorithms including demonstrated benefits when used with Soft
Actor-Critic for continuous control in MuJoCo \cite{rakelly2019efficient} and on-policy A2C in
MiniGrid and MiniPacMan \cite{goyal2019transfer}.

\section{Conclusion}

In this work we highlight two important differences between supervised learning and \gls{RL}:
First, the training data is generated using the learned model. Consequently, using stochastic
regularization methods can induce adverse effects and reduce the quality of the data.
We conjecture that this explains the observed lower performance of Batch Normalization and
Dropout.
Second, in \gls{RL}, we often encounter a noisy, low-data regime early on in
training, complicating the extraction of general features. 

We argue that these differences should inform the choice of regularization techniques used in RL.
To mitigate the adverse effects of stochastic regularization, we propose
\acrfull{SNI} which only selectively injects noise into the model, preventing reduced data quality and
higher gradient variance through a noisy critic.
On the other hand, to learn more compressed and general features in the noisy low-data regime, we propose \acrfull{IBAC},
which utilizes an variational information bottleneck as part of the agent.

We experimentally demonstrate that the \gls{VIB} is able to extract better features in the low-data
regime and that this translates to better generalization of \gls{IBAC} in \gls{RL}.
Furthermore, on complex environments, \gls{SNI} is key to good performance, allowing the combined
algorithm, \gls{IBAC}-\gls{SNI}, to achieve state of the art on challenging
generalization benchmarks.
We believe the results presented here can inform a range of future works, both to improve 
existing algorithms and to find new regularization techniques adapted to \gls{RL}.

\section*{Acknowledgments}
We would like to thank Shimon Whiteson for his helpful feedback,
Sebastian Lee, Luke Harris, Hiske Overweg and Patrick Fernandes for help with experimental evaluations and 
Adrian O'Grady, Jaroslaw Rzepecki and Andre Kramer for help with the computing infrastructure.
M. Igl is supported by the UK EPSRC CDT on Autonomous Intelligent Machines and Systems.

\bibliography{refs}
\bibliographystyle{plain}

\newpage

\appendix

\section{Dropout with SNI}
\label{sec:ap:dropout}

In order to apply \gls{SNI} to Dropout, we need to decide how to `suspend' the noise to compute
$\bar{\pi}_\theta$. 
While one could apply no dropout mask and scale the activations accordingly, we empirically found it
to be better to instead sample one dropout mask and keep it fixed for all gradient updates using the thus
collected data. This follows the implementation used in \citep{cobbe2018quantifying}.

\section{Supervised Classification Task}
\label{sec:ap:toy}

\begin{figure}[ht]
    \centering
    \begin{subfigure}{0.49\columnwidth}
        \includegraphics[width=\linewidth]{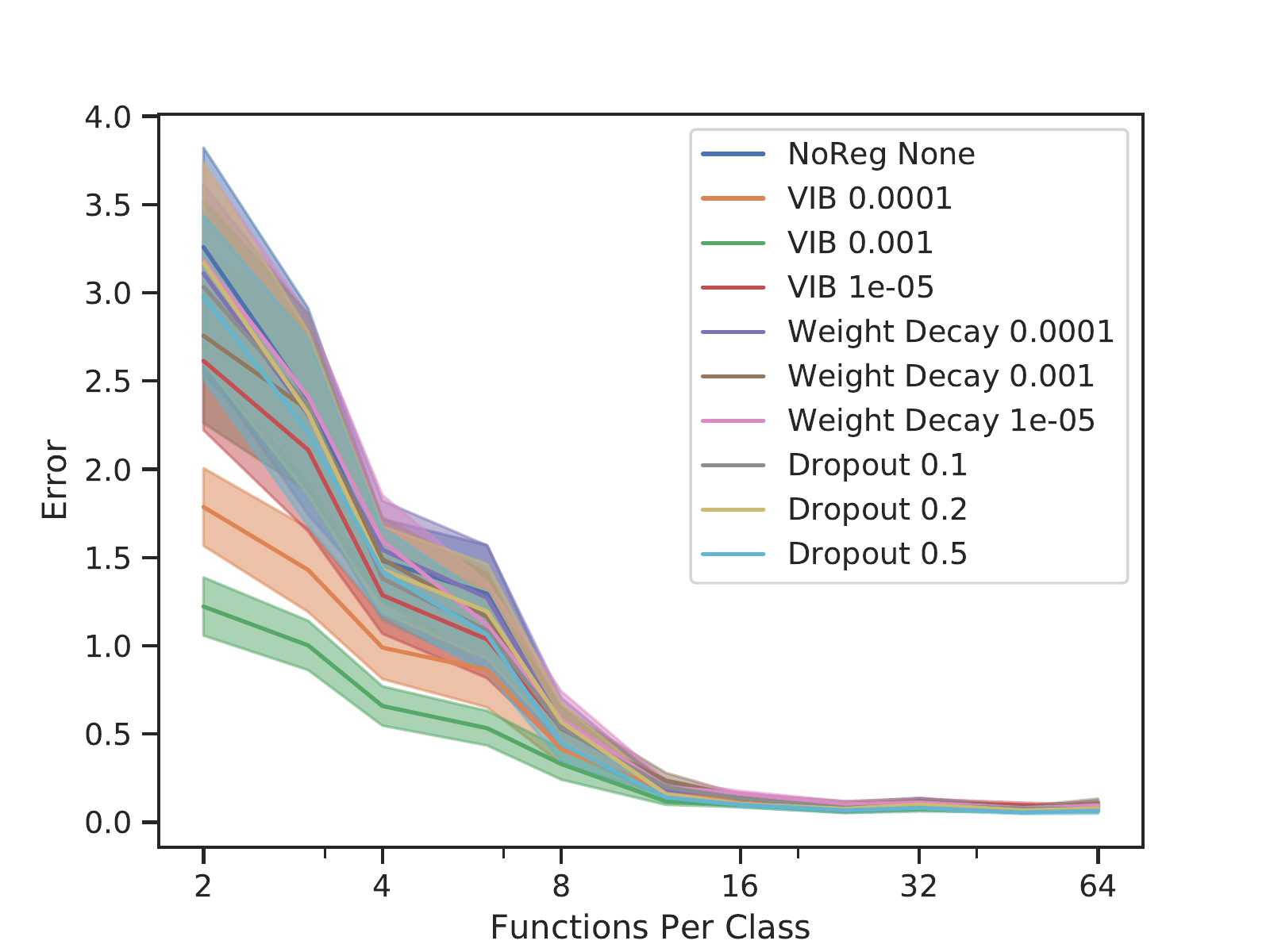}
    \end{subfigure}
    \begin{subfigure}{0.49\columnwidth}
        \includegraphics[width=\linewidth]{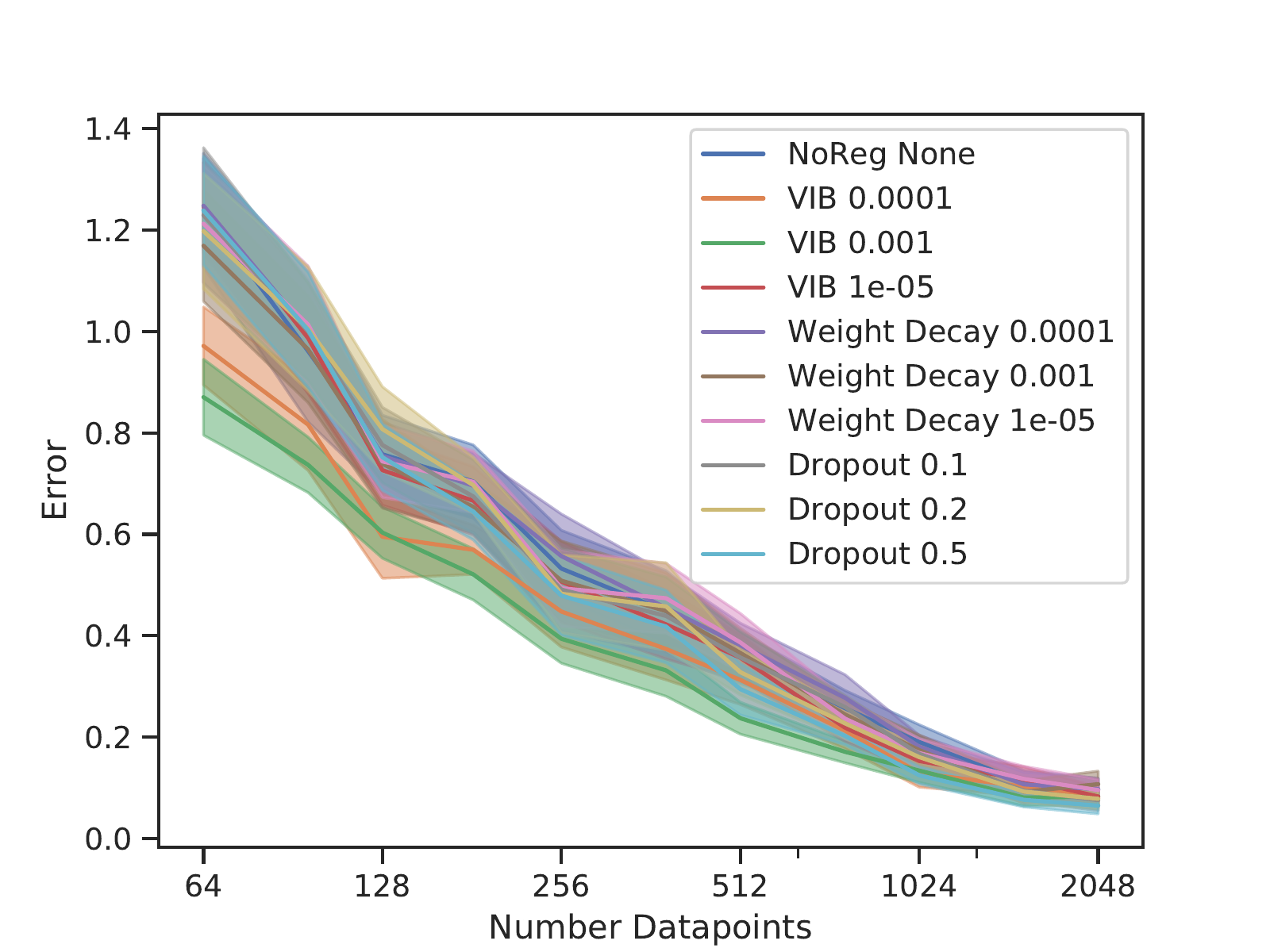}
    \end{subfigure}
    \caption{Loss function (error) on test set. Same results as in main text, but for multiple
    hyperparameters. The qualitative results are stable under a wide range of hyperparameters.
     }
    \label{fig:ap:toy}
\end{figure}

\paragraph{Network architecture and hyperparameters} The network consist of a 1D-convolutional layer
with 10 filters and a kernel size of 11 followed by two hidden, fully connected layers of size 1024
and 256 and the last layer which outputs $n_c$ logits. When the \gls{VIB} or Dropout are used, they
are applied to the last hidden layer. We use a learning rate of $1e-4$. The relative weight for
weight decay was $\lambda_w=1e-3$, which performed best out of $\{1e-2, 1e-3, 1e-4, 1e-5\}$. For the
\gls{VIB} we used $\beta=1e-3$, which performed best out of $\{1e-2, 1e-3, 1e-4, 1e-5\}$. Lastly,
For dropout we tested the dropout rates $p_d\in\{0.1, 0.2, 0.5\}$, out of which $0.2$ performed
best. Our results were stable across a range of hyperparameters, see \cref{fig:ap:toy}.

\begin{figure}[ht]
    \centering
    \includegraphics[width=\linewidth]{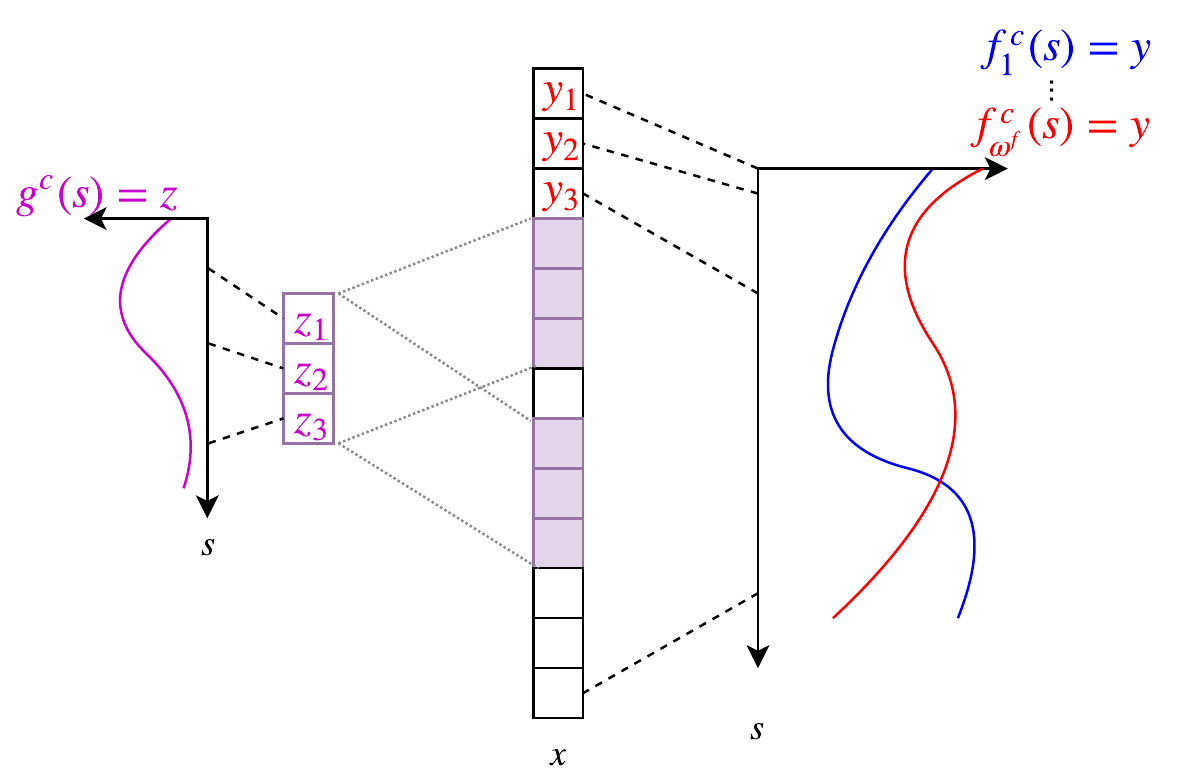}
    \caption{Generation of the input data $x$: We embed the information
    about $c$ twice, once through $f^c$ and once through $g^c$. See text for details.
    }
    \label{fig:ap:toy1}
\end{figure}

\paragraph{Data generation}
For each data-point, after drawing the class label $c_i$, we want to encode the information about
$c_i$ in two ways, using the encoding functions $f^c$ and $g^c$ which use one of $\omega^f$ and
$\omega^g$ different patterns to encode the information. The larger $\omega$ is, the less \emph{general}
the encoding is as it applies to fewer data-points.
Note that there are $\omega$ different patterns \emph{per class}.

We generate the patterns by first generating a set of random functions $\{f^c_j\}_{j=1}^{\omega^f}$
and $\{g^c_j\}_{j=1}^{\omega^g}$ by randomly drawing Fourier coefficients from $[0,1]$.
Those functions are converted into vectors by evaluating them at $d_x$ sorted points randomly drawn from
$[0,1]$.
The resulting pattern-vectors for $f^c$ will have a dimension of $d_x$, whereas the ones for $g^c$
will be smaller, $d_g<d_x$.

To encode the information about $c_i$ we first choose one pattern from $\{f^c_j\}_j$ (slightly
overloading notation between functions and pattern-vectors) and add some noise:
\begin{equation}
    x'_i=f^c_j + \epsilon_i \quad \text{where} \quad \epsilon \sim \mathcal{N}(0,\sigma_\epsilon) \quad \text{and} \quad j\sim \mathit{Cat}(\omega^f)
\end{equation}

Next, to also encode the information about $c_i$ using $g^c$, we choose one of the $\omega^g$
patterns $\{g^c_j\}_j$ and \emph{replace a part of the vector} $x'_i$, which is possible because the
$g^c$ patterns are shorter: $d_g<d_x$. The location of replacement is randomly drawn for each
data-point, but restricted to a a set of $n_g$ possible locations which are also random, but kept
fixed for the experiment and the same between training and testing set. The process is pictured in \cref{fig:ap:toy1}.

By changing the number of possible locations $n_g$ and the strength of the noise added to $f^c$,
$\sigma_\epsilon$,we can tune the relative difficulty of learning to recognize patterns $g^c$ and
$f^c$, allowing us to find a regime where both \emph{can} be found. Within this regime, our
qualitative results were stable. We use $n_g=3$ and $\sigma_\epsilon=1$. Furthermore, we have
for the dimension of of the observations $d_x=100$, and for the size of the patterns $g^c$ we have
$d_g=20$. We use $n_c=5$ different classes.

\section{Multiroom}

The observation space measures $11\times 11\times 3$ where the $3$ channels are used to encode
object type and object features like orientation or `open/closed' and `color' for doors on each of the $11\times 11$
spatial locations (see \cref{fig:grid} for a typical layout for $n_r=3$).

The agent uses a 3-layer CNN with $16, 32$ and $32$ filters respectively. All layers use a kernel of
size $2$. After the CNN, it uses one hidden layer of size 64 to which \gls{IBAC} or Dropout are applied if they are
used. Dropout uses $p_d=0.2$ and was tested for $\{0.1, 0.2, 0.5\}$. Both weight decay and
\gls{IBAC} were tried with a weighting factor of $\{1e-3, 1e-4, 1e-5, 1e-6\}$, with $1e-4$ performing
best for weight decay and $1e-6$ performing best for \gls{IBAC}. 
The output of the hidden layer is fed into a value function head and the policy
head.

We use a discount factor $\gamma=0.99$, a learning rate of $7e-4$, generalized value estimation
with$\lambda_{\text{GAE}}=0.95$ \citep{schulman2015high}, an entropy coefficient of
$\lambda_H=0.01$, value loss coefficient $\lambda_V=0.5$, gradient clipping at $0.5$ 
\citep{schulman2015high}, and \gls{PPO} with the Adam optimizer \citep{kingma2014adam}.

\section{Coinrun}
\label{sec:ap:coinrun}

\paragraph{Architecture and Hyperparameters} We use the same architecture ('Impala') and default policy
gradient hyperparameters as well as the codebase (\url{https://github.com/openai/coinrun}) from the
authors of \citep{cobbe2018quantifying} to ensure staying as closely as possible to their proposed
benchmark.

Dropout and \gls{IBAC} where applied to the last hidden layer and both, as well as weight decay,
were tried with the same set of hyperparameters as in Multiroom.
The best performance was achieved with $p_d=0.2$ for Dropout and $1e-4$ for \gls{IBAC} and weight
decay.
Batch normalization was applied between the layers of the convolutional part of the network.
Note that the original architecture in \citep{cobbe2018quantifying} uses Dropout also on earlier
layers, however, we achieve higher performance with our implementation.

\begin{figure}[ht]
    \begin{subfigure}[b]{0.49\columnwidth}
    \includegraphics[width=\linewidth]{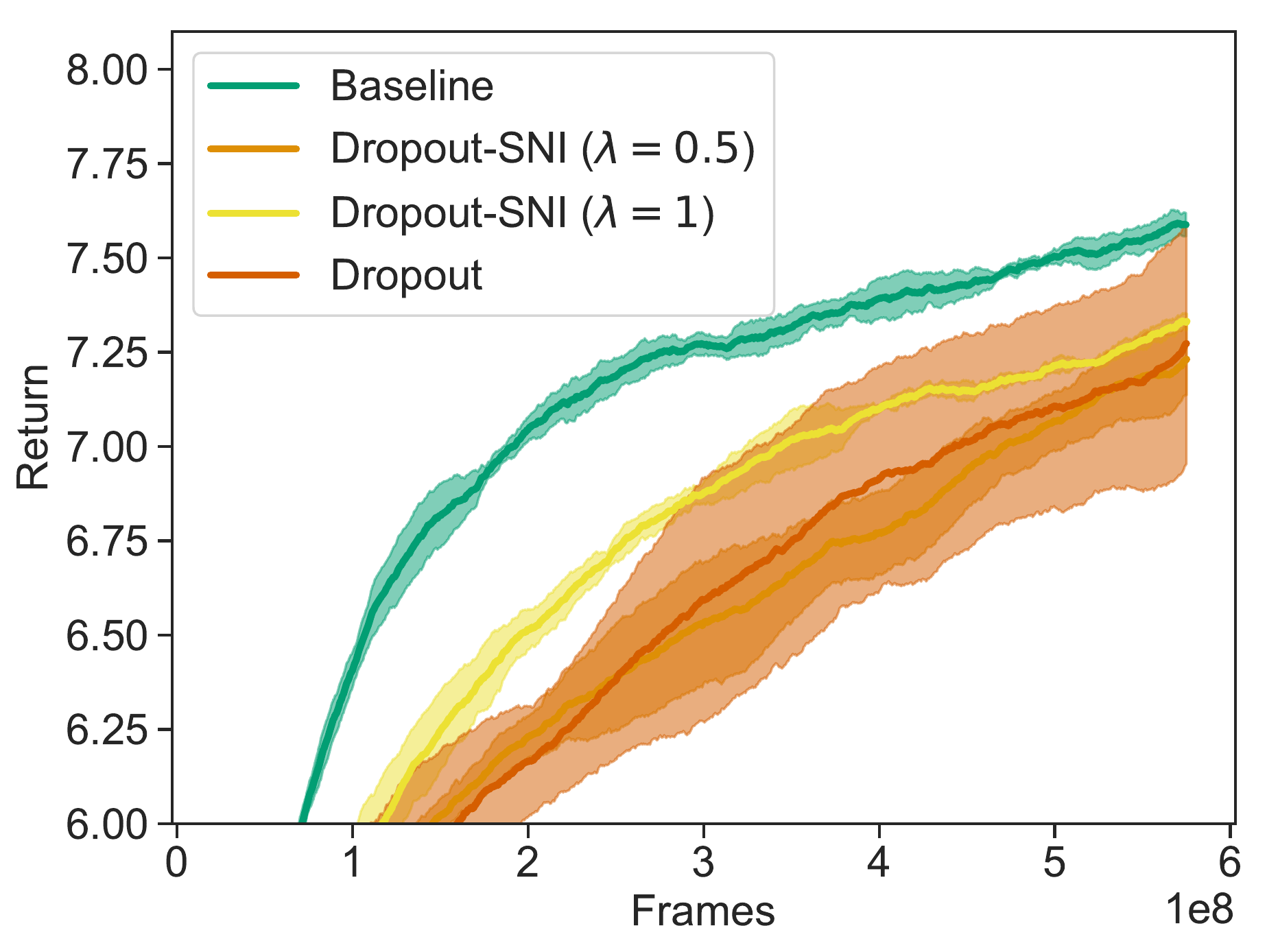}
    \end{subfigure}
    \begin{subfigure}[b]{0.49\linewidth}
    \includegraphics[width=\linewidth]{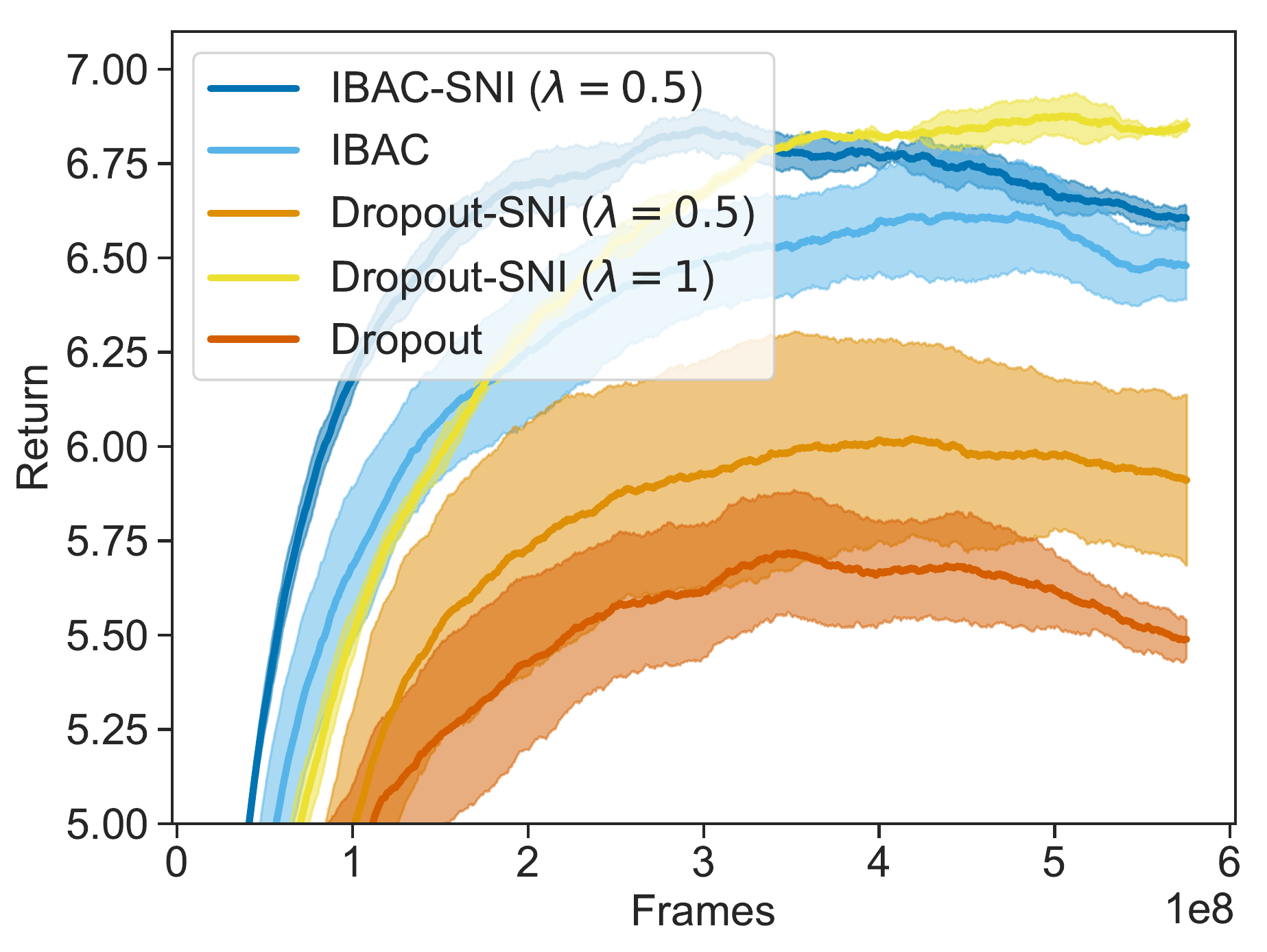}
    \end{subfigure}
    \centering  
    \caption{\emph{Left:} Comparison for different implementations of Dropout on the test
    environments. 
    \emph{Right:} Comparison of \gls{IBAC} and Dropout, with and without \gls{SNI},
    \emph{without weight decay and data augmentation}.
     }
    \label{fig:ap:coinrun2}
\end{figure}

In \cref{fig:ap:coinrun2} (left) we show results for Dropout with and without \gls{SNI} and for $\lambda=1$
and $\lambda=0.5$. We find that $\lambda=1$ learns fastest, possible due to the high importance
weight variance in the stochastic term in \gls{SNI} for $\lambda<1$ (see \cref{fig:coinrun}
(right)). However, all Dropout implementations converge to roughly the same value, significantly
below the `baseline' agent, indicating that Dropout is not suitable for combination with weight
decay and data augmentation.

In \cref{fig:ap:coinrun2} (right) we show the test performance for \gls{IBAC} and Dropout with and without
\gls{SNI}, \emph{without} using weight decay and data-augmentation. Again, we can see that \gls{SNI}
helps the performance. Interestingly, we can see that \gls{IBAC} does not prevent overfitting by
itself (one can see the performance decreasing for longer training) but does lead to faster
learning. 
Our conjecture is that it finds more general features early on in training, but ultimately overfits
to the test-set of environments without additional regularization.
This further indicates that
it's regularization is different to techniques such as weight decay, explaining why their
combination synergizes well.

\begin{figure}[ht]
    \centering  
    \begin{subfigure}[b]{0.49\columnwidth}
    \includegraphics[width=\linewidth]{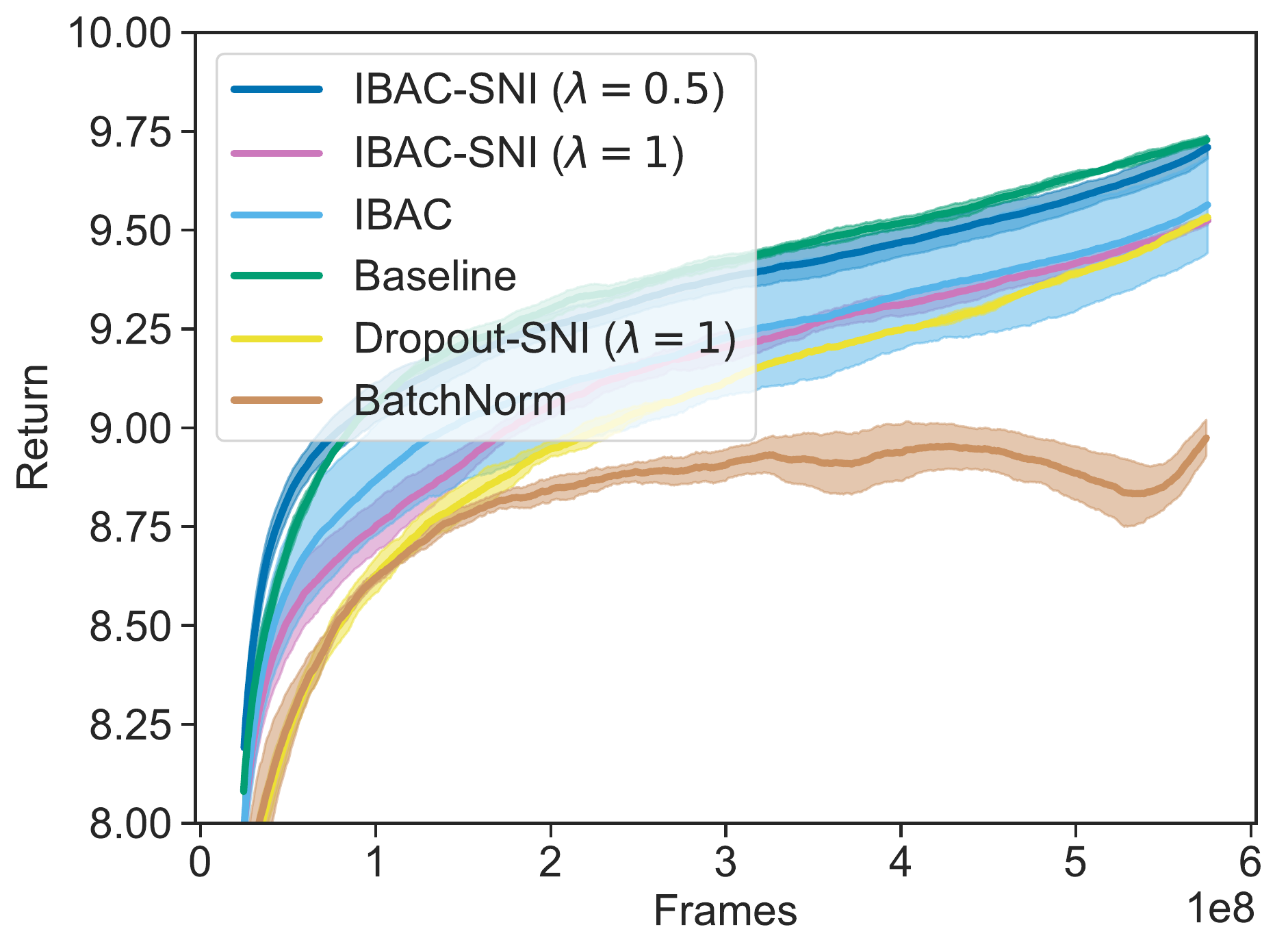}
    \end{subfigure}
    \begin{subfigure}[b]{0.49\linewidth}
    \includegraphics[width=\linewidth]{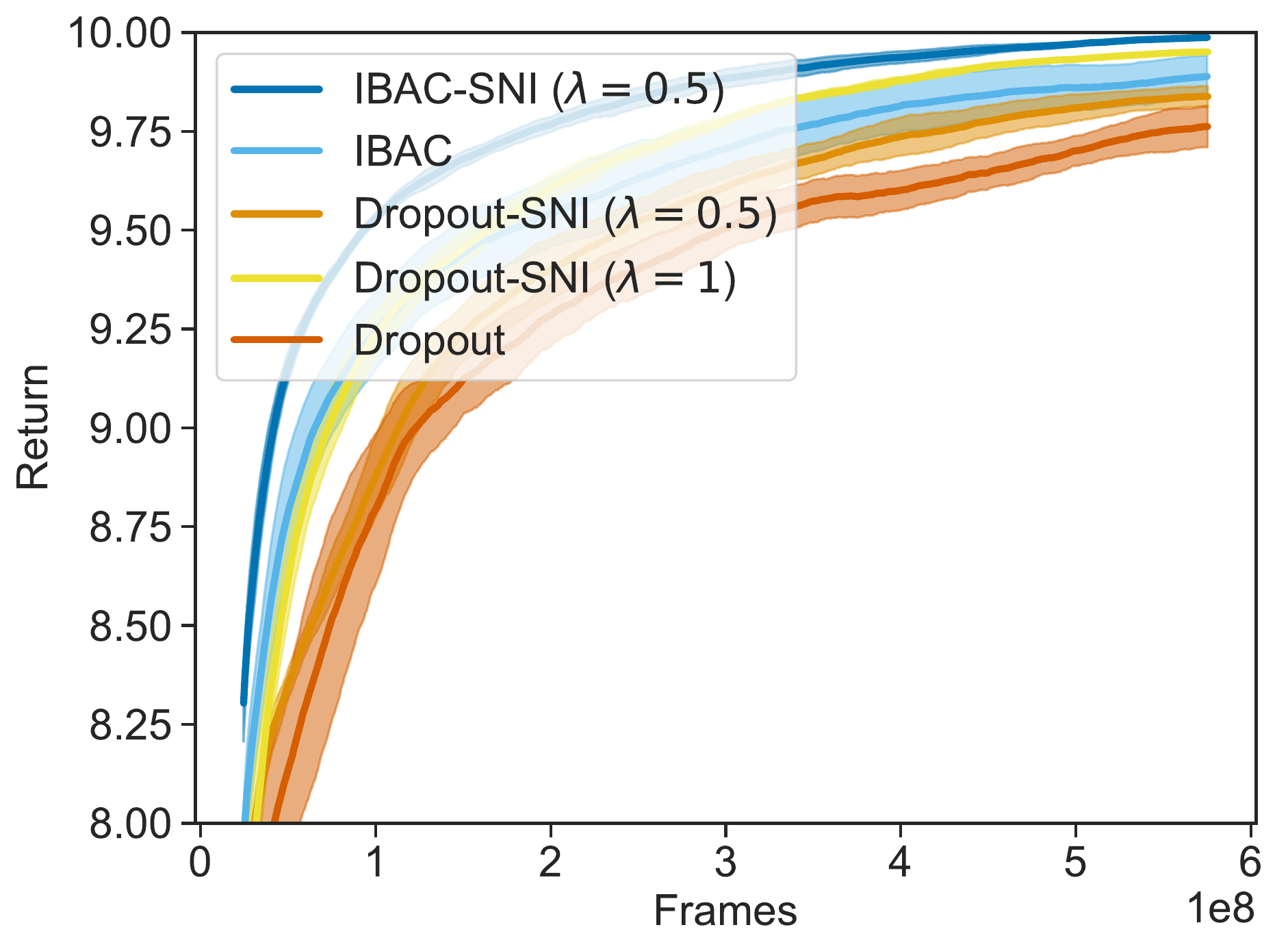}
    \end{subfigure}
    \caption{Training Performance with weight decay and data augmentation (left) and without (right)}
    \label{fig:ap:coinrun1}
\end{figure}

In \cref{fig:ap:coinrun1} we show the training set performance of our experiments.

\end{document}